\pdfoutput=1
\documentclass[nohyperref]{article}

\usepackage{microtype}
\usepackage{graphicx}
\usepackage{booktabs} % for professional tables

\usepackage[hyphens]{url}
\usepackage{hyperref}

\usepackage[accepted]{icml2022}

\usepackage{amsmath}
\usepackage{amssymb}
\usepackage{mathtools}
\usepackage{amsthm}

\usepackage[capitalize,noabbrev]{cleveref}

\theoremstyle{plain}
\newtheorem{theorem}{Theorem}[section]
\newtheorem{proposition}[theorem]{Proposition}

\theoremstyle{definition}

\theoremstyle{remark}

\usepackage[textsize=tiny]{todonotes}
\usepackage{subfig}
\usepackage{multirow}
\icmltitlerunning{Comprehensive Analysis of Negative Sampling in Knowledge Graph Representation Learning}

\begin{document}

\twocolumn[
\icmltitle{Comprehensive Analysis of Negative Sampling \\ in Knowledge Graph Representation Learning}

\icmlsetsymbol{equal}{*}

\begin{icmlauthorlist}
\icmlauthor{Hidetaka Kamigaito}{naist}
\icmlauthor{Katsuhiko Hayashi}{hokkaido}
%\icmlauthor{Firstname3 Lastname3}{comp}
%\icmlauthor{Firstname4 Lastname4}{sch}
%\icmlauthor{Firstname5 Lastname5}{yyy}
%\icmlauthor{Firstname6 Lastname6}{sch,yyy,comp}
%\icmlauthor{Firstname7 Lastname7}{comp}
%\icmlauthor{Firstname8 Lastname8}{sch}
%\icmlauthor{Firstname8 Lastname8}{yyy,comp}
\end{icmlauthorlist}

\icmlaffiliation{naist}{Nara Institute of Science and Technology (NAIST), Nara, Japan}
\icmlaffiliation{hokkaido}{Hokkaido University, Hokkaido, Japan}
%\icmlaffiliation{sch}{School of ZZZ, Institute of WWW, Location, Country}

\icmlcorrespondingauthor{Hidetaka Kamigaito}{kamigaito.h@is.naist.jp}
%\icmlcorrespondingauthor{Firstname2 Lastname2}{first2.last2@www.uk}

\icmlkeywords{Knowledge Graph Embedding, KGE, Natural Language Processing, NLP}

\vskip 0.3in
]

\printAffiliationsAndNotice{} % leave blank if no need to mention equal contribution

\begin{abstract}
Negative sampling~(NS) loss plays an important role in learning knowledge graph embedding~(KGE) to handle a huge number of entities. However, the performance of KGE degrades without hyperparameters such as the margin term and number of negative samples in NS loss being appropriately selected. Currently, empirical hyperparameter tuning addresses this problem at the cost of computational time. To solve this problem, we theoretically analyzed NS loss to assist hyperparameter tuning and understand the better use of the NS loss in KGE learning. Our theoretical analysis showed that scoring methods with restricted value ranges, such as TransE and RotatE, require appropriate adjustment of the margin term or the number of negative samples different from those without restricted value ranges, such as RESCAL, ComplEx, and DistMult. We also propose subsampling methods specialized for the NS loss in KGE studied from a theoretical aspect. Our empirical analysis on the FB15k-237, WN18RR, and YAGO3-10 datasets showed that the results of actually trained models agree with our theoretical findings.
\end{abstract}

\section{Introduction}
\label{sec:intro}

Knowledge graph completion~(KGC) has attracted increasing attention as a research topic for automatically inferring new links in a KG that are likely but not yet known to be true. KG embedding~(KGE) methods have been actively pursued as a scalable approach to KGC~\cite{transe,distmult,complex,rotate}. Since a KG contains a huge number of entities, negative sampling~(NS) loss~\cite{ns} is widely used in KGE training from the viewpoint of computational efficiency. However, NS loss was originally proposed for word-representation learning~\cite{ns}, and there are several issues that need to be considered when using NS loss for training KGE models.

While the original NS loss was designed to be used with scoring methods on the basis of the inner product, which have no limit on the value range, many KGE models such as TransE~\cite{transe} and RotatE~\cite{rotate} use score functions with value ranges limited by the $p$-norm distance. To apply NS loss to distance-based KGE models, a margin term has been introduced as a heuristic means of adjusting the value ranges of the score functions~\cite{transe,rotate}. This margin term is not present in the original NS loss, and there has been no discussion of how it affects the KGE model training. In certain KGE studies~\cite{complex,rotate}, the NS loss is normalized by the number of negative samples. This normalization term also does not exist in the original NS loss, and the validity of the term has not yet been verified.

In KG representation learning, self-adversarial NS (SANS)~\cite{rotate} is often used to reduce the effect of negative samples that do not carry meaningful information to the learning process. To generate negative samples, SANS uses probability distributions calculated from the current embedding model instead of using the uniform noise distribution. However, the normalization term of the probability distribution is computationally expensive and is approximated in practice by randomly sampled examples. Though certain characteristics of SANS have been investigated~\cite{olddog,unify}, more detailed analysis is needed because the previous studies did not take into account the effect of the approximation of the normalization term.

Due to the difference from the original NS loss previously described, we have to consider various combinations of score functions, loss functions, and hyperparameters in KGE training. It is known that KGE performance can be significantly improved by selecting the appropriate combinations~\cite{olddog,benchmarking,zhang-etal-2022-efficient}, but it is unclear what combinations are theoretically appropriate. Therefore, the current state-of-the-art method relies on empirical hyperparameter tuning, which requires a huge amount of computation time to consider various combinations.

The nature of KG data also makes the NS loss in KGE unique. Since a triplet, which is a training instance of KGE, appears at most once in a KG, KGE learning suffers from the serious data-sparseness problem. Subsampling has been developed as a method for mitigating the data-sparseness problem in word-representation learning. However, there has been very little discussion on subsampling in KGE learning.

To address the aforementioned issues, we theoretically analyzed the NS loss to assist hyperparameter tuning and understand the better use of the NS loss in KGE learning. Our theoretical analysis showed that scoring methods with restricted value ranges, such as TransE, RotatE, require appropriate adjustment of the margin term or number of negative samples different from those without restricted value ranges, such as RESCAL, ComplEx, and DistMult. We present novel subsampling methods specialized for the NS loss in KGE learning from a theoretical perspective. Our empirical analysis on the FB15k-237, WN18RR, and YAGO3-10 datasets showed that the results of the models trained on the real-world datasets agree with our theoretical findings. Our code is available at \url{https://github.com/kamigaito/icml2022}.

\section{NS Loss in KGE}

In this section, we formulate KGE and explain the difference between the original NS loss~\citep{ns} and that used in KGE~\citep{rotate}.

\subsection{Formulation of KGE}

We denote a triplet representing entities $e_{i}$, $e_{j}$ and their relation $r_{k}$ as $(e_{i},r_{k},e_{j})$. In a typical KGC task, the model receives a query $(e_{i},r_{k},?)$ or $(?,r_{k},e_{j})$ and predicts the entity corresponding to $?$. Let the input query be $x$ and its answer entity be $y$. The probability $p_{\theta}(y|x)$ that $y$ is predicted from $x$ under the score function $s_{\theta}(x,y)$ on the basis of the model parameter $\theta$ is defined as follows using the softmax function:
\begin{equation}
 p_{\theta}(y|x)=\frac{
\exp{(s_{\theta}(x,y))}
}{
\sum_{y'\in Y}{
\exp{(s_{\theta}(x,y'))}
}
},
\label{eq:softmax}
\end{equation}
where $Y$ is the set of entities and the size $|Y|$ can be large in KGE learning.

\subsection{NS Loss Functions}

The softmax function can explicitly represent probabilities, but the normalization term increases the learning time. For this reason, the original NS loss was proposed as a method to approximate the softmax function without normalization during training~\cite{ns}. For observables $D=\{(x_{1},y_{1}),\cdots, (x_{n},y_{n})\}$ that follow $p_{d}(x,y)$, the original NS loss is defined as
\begin{align}
 -&\frac{1}{|D|}\sum_{(x,y) \in D} \Bigl[\log(\sigma(s_{\theta}(x,y))) \nonumber\\
 &+ \sum_{y_{i}\sim p_n(y_{i}|x)}^{\nu}\log(\sigma(-s_{\theta}(x,y_i)))\Bigr],\label{eq:ns:loss}
\end{align}
where $p_n(y|x)$ is the noise distribution, $\sigma$ is the sigmoid function, and $\nu$ is the number of negative samples per positive sample $(x,y)$.

KGE uses the following NS loss~\cite{rotate,ahrabian-etal-2020-structure} with the addition of the margin term $\gamma$ and normalization term for $\nu$ to Eq.~(\ref{eq:ns:loss}):
\begin{align}
 -&\frac{1}{|D|}\sum_{(x,y) \in D} \Bigl[\log(\sigma(s_{\theta}(x,y)+\gamma)) \nonumber\\
 &+ \frac{1}{\nu}\sum_{y_{i}\sim p_n(y_{i}|x)}^{\nu}\log(\sigma(-s_{\theta}(x,y_i)-\gamma))\Bigr].\label{eq:ns:loss:kge}
\end{align}

Hereafter, we call Eq.~(\ref{eq:softmax}) when the loss functions, such as Eqs.~(\ref{eq:ns:loss}) and~(\ref{eq:ns:loss:kge}), are minimized, which means $s_{\theta}(x,y)$ reaches an optimal solution, as the \textit{objective distribution}. In the next section, we discuss the difference between the two loss functions.

\section{Theoretical Analysis}
\label{sec:theory}

\subsection{Equivalence of the Two Loss Functions}

By comparing Eq.~(\ref{eq:ns:loss}) with Eq.~(\ref{eq:ns:loss:kge}), we can show the following proposition.
\begin{proposition}
Eqs.~(\ref{eq:ns:loss}) and~(\ref{eq:ns:loss:kge}) have the same objective distribution:
\begin{equation}
    \frac{p_d(y|x)/p_n(y|x)}{\sum_{y'\in Y}{(p_d(y'|x)/p_n(y'|x))}}.
\end{equation}
\label{prop:obj}
\end{proposition}

See Appendix~\ref{app:subsec:obj} for the proof.
From Prop.~\ref{prop:obj}, we can see that the existence of $\nu$ and $\gamma$ has no effect on the distribution that the model will fit when the NS loss reaches the optimal solution.
However, if the score function does not have the value range $(-\infty,+\infty)$, the score limitation prevents a model from reaching the optimal solution, and we need to discuss this point.

\subsection{Roles of the Margin Term $\gamma$}
\label{subsec:margin}

In this section, we discuss the role of the margin term $\gamma$ from a theoretical perspective.

\subsubsection{Value ranges of scoring methods}
\label{subsubsec:margin:value}

Distance-based scoring methods, such as TransE~\cite{transe} and RotatE~\cite{rotate}, have score functions with a restricted value range. Distance-based scoring methods are generally expressed using the $p$-norm as follows:
\begin{equation}
 -||f_{\theta}(x, y)||_{p},
 \label{eq:dist}
\end{equation}
where $f_{\theta}(x, y)$ is a function that returns a vector value for $(x,y)$. From Eq.~(\ref{eq:dist}), when $f_{\theta}(x, y)$ can represent any vector, the value range of a distance-based scoring method is $(-\infty,0]$. Due to the value range limitation, the following proposition holds for distance-based scoring methods.
\begin{proposition}
In Eq.~(\ref{eq:ns:loss:kge}), the distance-based scoring function cannot reach the optimal solution when there exists $(x,y)$ that satisfies $\exp(\gamma)p_{n}(y|x)<p_{d}(y|x)$.
\label{prop:bound1}
\end{proposition}
See Appendix~\ref{app:subsec:bound} for the proof.
Since the value range restriction causes this problem in Eq.~(\ref{eq:ns:loss:kge}), the smaller $\exp(\gamma)p_{n}(y|x)$ than $p_{d}(y|x)$ is, the less able to approach the optimal solution the distance-based scoring function is.
It is clear from Prop.~\ref{prop:bound1} that we need to choose the noise distribution and $\gamma$ appropriately when learning a distance-based scoring method with the NS loss in Eq.~(\ref{eq:ns:loss:kge}). 
To satisfy $\exp(\gamma)p_{n}(y|x)\geqq p_{d}(y|x)$ in the NS loss in Eq.~(\ref{eq:ns:loss:kge}), we can simply set a sufficiently large $\gamma$. Since uniform distribution is generally used for $p_{n}(y|x)$ in KGE and $p_{d}(y|x)$ does not exceed $1$ from the definition of probability, $\exp(\gamma)$ should be larger than the number of labels $|Y|$. To make a distance-based method capable to reach the optimal solution, it is appropriate to use $\gamma$ satisfying
\begin{equation}
\gamma \geqq \log(|Y|).
\label{eq:margin:condition}
\end{equation}
Note that scoring methods with unlimited value ranges, such as RESCAL~\cite{rescal}, ComplEx~\citep{complex}, and DistMult~\citep{distmult} are not related to the discussion of Prop.~\ref{prop:bound1} and Eq.~(\ref{eq:margin:condition}) that indicate the importance of adjusting $\gamma$.

\subsubsection{Gradient changes}

From the discussion in \S \ref{subsubsec:margin:value}, it seems that setting a large value for $\gamma$ will facilitate the learning of the model without facing any disadvantages. However, the following proposition shows that $\gamma$ cannot be set freely.
\begin{proposition}
The margin term $\gamma$ affects the gradient of the NS loss in Eq.~(\ref{eq:ns:loss:kge}).
\label{prop:grad:gamma}
\end{proposition}

See Appendix~\ref{app:subsec:grad:gamma} for the proof.
From Prop.~\ref{prop:grad:gamma}, we can see that when we change $\gamma$, we also have to set the other hyperparameters related to the gradient appropriately. Thus, when using a distance-based scoring method, we need to set a sufficiently large $\gamma$ and adjust the hyperparameters related to the gradient appropriately.

\subsection{Roles of the Number of Negative Samples $\nu$}
\label{subsec:nss}

Next, we discuss the effect of $\nu$ on learning from a theoretical perspective.

\subsubsection{Value ranges of scoring methods}

Similar to Prop.~\ref{prop:bound1}, the following proposition holds:

\begin{proposition}
In Eq.~(\ref{eq:ns:loss}), the distance-based scoring function cannot reach the optimal solution when there exists $(x,y)$ that satisfies $\nu p_{n}(y|x)<p_{d}(y|x)$.
\label{prop:bound2}
\end{proposition}
See Appendix~\ref{app:subsec:bound} for the proof.
Similar to Prop.~\ref{prop:bound1}, in Eq.~(\ref{eq:ns:loss}), the smaller $\nu p_{n}(y|x)$ than $p_{d}(y|x)$ is, the less able to approach the optimal solution the distance-based scoring function is.
From Prop.~\ref{prop:bound2}, a distance-based scoring method should satisfy the condition $\nu p_{n}(y|x) \geqq p_{d}(y|x)$ to reach the optimal solution in the NS loss of Eq.~(\ref{eq:ns:loss}).
When using a uniform distribution for $p_{n}(y|x)$, then
\begin{equation}
\nu \geqq |Y|,
\label{eq:nu:condition}
\end{equation}
must be satisfied.
However, since the goal of the NS loss is to reduce the computational cost by setting $\nu<|Y|$, this condition would greatly degrade the advantage of the NS loss in Eq.~(\ref{eq:ns:loss}).
Note that similar to $\gamma$, the discussion of Prop.~\ref{prop:bound2} and Eq.~(\ref{eq:nu:condition}) is not related to scoring methods that have no limited value range.

\subsubsection{Gradient changes}

Furthermore, we can induce the following proposition by the Monte Carlo method on the basis of the law of large numbers.
\begin{proposition}
When $\nu$ is sufficiently large, $\nu$ affects the gradient of the NS loss in Eq.~(\ref{eq:ns:loss}), whereas $\nu$ does not affect the gradient of the NS loss in Eq.~(\ref{eq:ns:loss:kge}).
\label{prop:ns}
\end{proposition}

See Appendix~\ref{app:subsec:ns} for the proof.
When using the NS loss in Eq.~(\ref{eq:ns:loss:kge}), if $\nu$ is sufficiently large, it is unnecessary to tune $\nu$ in detail.
However, for the NS loss in Eq.~(\ref{eq:ns:loss}), when increasing $\nu$ for distance-based scoring methods, it is necessary to adjust hyperparameters related to the gradient.

\subsection{Effects of Noise Distributions}

Instead of uniform noise distributions, SANS presented by~\citet{rotate} uses the output of the training model as the noise distribution for sampling negative entities. Though it is difficult to apply the theoretical analysis conducted on the loss functions in Eqs.~(\ref{eq:ns:loss}) and~(\ref{eq:ns:loss:kge}) directly to SANS, the Monte Carlo method on the basis of the law of large numbers enables us to derive the following proposition.
\begin{proposition}
When $\nu$ is sufficiently large, SANS becomes equivalent to the NS loss set to $p_{n}(y|x)=p_{\theta}(y|x)$ in Eq.~(\ref{eq:ns:loss:kge}).
\label{prop:sans}
\end{proposition}
See Appendix~\ref{app:subsec:sans} for the proof. Prop.~\ref{prop:sans} means that if $\nu$ is sufficiently large, SANS can be treated as the NS loss expressed in the same form as in Eq.~(\ref{eq:ns:loss:kge}). This enables us to apply to SANS to what we have discussed for Eq.~(\ref{eq:ns:loss:kge}).

\subsection{Subsampling for KGE}
\label{subsec:subsamp}

The discussion thus far has been on the assumption that the NS loss function fits the model to the distribution $p_{d}(y|x)$ defined from the observed data. However, what the NS loss actually does is to fit the model to the true distribution $p'_{d}(y|x)$ that exists behind the observed data.
To fill in the gap between $p_{d}(y|x)$ and $p'_{d}(y|x)$, we reformulate the NS loss in Eq.~(\ref{eq:ns:loss:kge}) by the Monte Carlo method as
\begin{align}
(\ref{eq:ns:loss:kge})\approx-&\sum_{x,y} \Bigl[\log(\sigma(s_{\theta}(x,y)+\gamma))p_{d}(x,y) \nonumber\\
&+p_n(y|x)\log(\sigma(-s_{\theta}(x,y)-\gamma))p_{d}(x)\Bigr],
\label{eq:subsamp:intro}
\end{align}
and consider replacements of $p_{d}(x,y)$ with $p'_{d}(x,y)$ and $p_{d}(x)$ with $p'_{d}(x)$.
By assuming two functions, $A(x,y)$ and $B(x)$, that convert $p_{d}(x,y)$ into $p'_{d}(x,y)$ and $p_{d}(x)$ into $p'_{d}(x)$, we further reformulate Eq.~(\ref{eq:subsamp:intro}) as follows.
\begin{align}
& -\sum_{x,y} \Bigl[\log(\sigma(s_{\theta}(x,y)+\gamma))p'_{d}(x,y) \nonumber\\
&+p_n(y|x)\log(\sigma(-s_{\theta}(x,y)-\gamma))p'_{d}(x)\Bigr]\nonumber\\
=&-\sum_{x,y} \Bigl[\log(\sigma(s_{\theta}(x,y)+\gamma))A(x,y)p_{d}(x,y) \nonumber\\
&+p_n(y|x)\log(\sigma(-s_{\theta}(x,y)-\gamma))B(x)p_{d}(x)\Bigr]\nonumber\\
\approx& -\frac{1}{|D|}\sum_{(x,y) \in D} \Bigl[A(x,y)\log(\sigma(s_{\theta}(x,y)+\gamma)) \nonumber\\
&+\frac{1}{\nu}\sum_{y_{i}\sim p_n(y_{i}|x)}^{\nu}B(x)\log(\sigma(-s_{\theta}(x,y_i)-\gamma))\Bigr].
\label{eq:subsamp}
\end{align}
See Appendix \ref{app:derivation} for the detailed derivation.
\citet{ns} proposed subsampling for the NS loss in Eq.~(\ref{eq:ns:loss}) to balance the appearance probability of words and word pairs by discounting their frequency.
In Eq.~(\ref{eq:ns:loss}), we can consider a word as $x$ and a word pair as $(x,y)$. We denote the appearance probability of $x$ as $p_d(x)$ and $(x,y)$ as $p_d(x,y)$ in Eq.~(\ref{eq:subsamp}). Since $A(x,y)$ and $B(x)$ adjust $p_d(x,y)$ and $p_d(x)$ in Eq.~(\ref{eq:subsamp}), we can understand that subsampling has the same role with $A(x,y)$ and $B(x)$.

To the best of our knowledge, no paper discusses the use of subsampling in KGE.
However, \citet{rotate} used subsampling which follows word2vec in their implementation\footnote{\url{https://github.com/DeepGraphLearning/KnowledgeGraphEmbedding}}. Regarding $B(x)$ as $B(x,y)$, we can understand that their subsampling fills the gap between $p_{d}(y|x)$ and $p'_{d}(y|x)$ in Eq.~(\ref{eq:subsamp}) as follows:
\begin{equation}
 A(x,y)=B(x,y)=\frac{\frac{1}{\sqrt{\#(x,y)}}|D|}{\sum_{(x',y') \in D}\frac{1}{\sqrt{\#(x',y')}}}.
 \label{eq:subsamp:default}
\end{equation}
Here, $\#$ is the symbol for frequency, and $\#(x,y)$ represents the frequency of $(x,y)$. Note that the actual $(x,y)$ occurs at most once in the KG, so when $(x,y)=(e_{i},r_{k},e_{j})$, they approximate the frequency of $(x,y)$ as follows:
\begin{equation}
 \#(x,y) \approx \#(e_{i},r_{k})+\#(r_{k},e_{j}).
 \label{eq:subsamp:approx}
\end{equation}
We can interpret this approximation as the backoff~\cite{katz1987estimation} of $\#(x,y)$ to $\#(e_{i},r_{k})$ and $\#(r_{k},e_{j})$.

We derived our proposed KGE-specific subsampling methods from a theoretical perspective on the basis of Eq.~(\ref{eq:subsamp}) that discounts by frequency as in Eq.~(\ref{eq:subsamp:default}). This derivation depends on how we make assumptions about $p'_{d}(y|x)$. First, the derivation is based on the assumption that in $p'_{d}(y|x)$, $(x,y)$ originally has a frequency, but the observed one is at most 1. In this case, we cannot actually calculate the frequency of $\#(x,y)$, as in Eq.~(\ref{eq:subsamp:default}). Thus, we need to use the approximation in Eq.~(\ref{eq:subsamp:approx}). 
Since $A(x,y)$ needs to discount the frequency of $(x,y)$, and $B(x)$ needs to discount that of $x$, we can derive the following subsampling method:
\begin{eqnarray}
 A(x,y)&=&\frac{\frac{1}{\sqrt{\#(x,y)}}|D|}{\sum_{(x',y') \in D}\frac{1}{\sqrt{\#(x',y')}}}, \nonumber\\
 B(x)&=&\frac{\frac{1}{\sqrt{\#x}}|D|}{\sum_{x' \in D}\frac{1}{\sqrt{\#x'}}}.
 \label{eq:subsamp:freq}
\end{eqnarray}

In $p'_{d}(y|x)$, however, if we assume that $(x,y)$ has frequency $1$ at most, as in the observation, then $p'_d(y|x)=p'_d(x,y)/p'_d(x) \propto 1/p'_d(x)$, so $p'_d(y|x)$ is the same for an $x$ independent from $y$. Therefore, under this assumption, we only consider a discount for $p_d(x)$ and derive the following subsampling method:
\begin{equation}
 A(x,y)=B(x)=\frac{\frac{1}{\sqrt{\#x}}|D|}{\sum_{x' \in D}\frac{1}{\sqrt{\#x'}}}.
 \label{eq:subsamp:uniq}
\end{equation}

Although we derive our methods using Eqs.~(\ref{eq:subsamp:freq}) and~(\ref{eq:subsamp:uniq}) from a theoretical perspective, $p'_{d}(y|x)$, which is the target of generalization in the actual task, varies depending on the dataset. Therefore, we cannot discuss the superiority or inferiority of our subsampling methods only from a theoretical perspective, and it is necessary to verify the methods using the development data in actual use.

\begin{figure*}[t!]
\centering
\includegraphics[width=\textwidth]{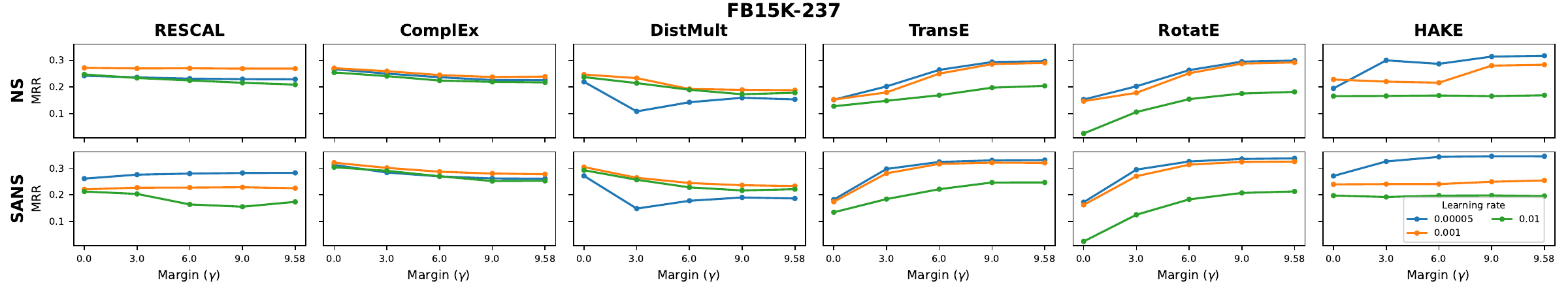}
\includegraphics[width=\textwidth]{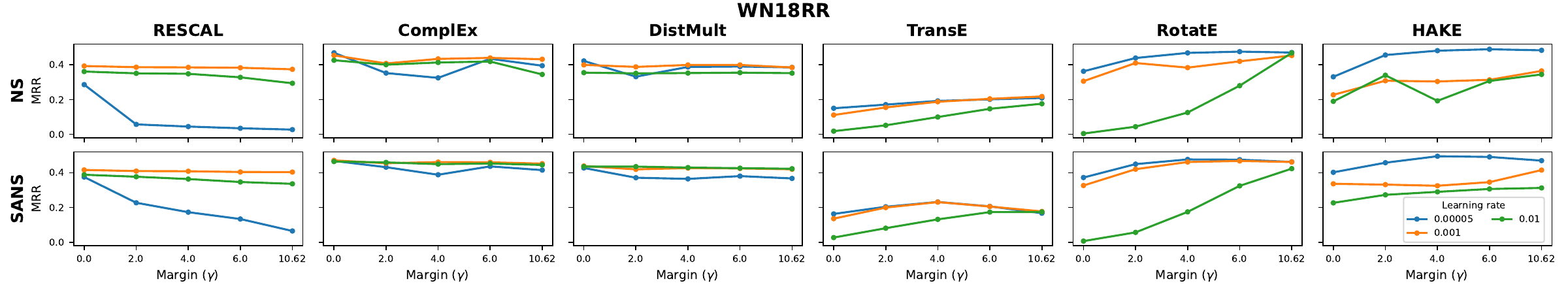}
\caption{MRR scores for each margin term $\gamma$ in each learning rate on FB15k-237 and WN18RR.\label{fig:margin:zero}}
\end{figure*}

\section{Empirical Analysis}
We examined whether the theoretical analysis discussed in \S \ref{sec:theory} is valid for actual datasets and models.
\subsection{Settings}
\label{subsec:analysis:settings}

\paragraph{Datasets}
We used FB15k-237 \citep{fb15k-237}, WN18RR, and YAGO3-10 \citep{conve} as the datasets. Table \ref{tab:data:stats} shows the statistics for each dataset.

\begin{table}[t]
\caption{Statistics for each dataset.\label{tab:data:stats}}
    \centering
    \resizebox{0.95\columnwidth}{!}{
    \small
    \begin{tabular}{llllll}
    \toprule
    \multirow{2}{*}{\textbf{Dataset}}&\multirow{2}{*}{\textbf{Entities}}&\multirow{2}{*}{\textbf{Relations}}&\multicolumn{3}{c}{\textbf{Tuples}}\\
    \cmidrule(lr){4-6}
 &  &  & \textbf{Train} & \textbf{Valid} & \textbf{Test}\\
    \midrule
FB15k-237 &14,541 &237 &272,115 &17,535& 20,466  \\
WN18RR &40,943 &11 &86,835 &3,034 &3,134\\
YAGO3-10 & 123,182 & 37 & 1,079,040 & 4,978 & 4,982 \\
\bottomrule
    \end{tabular}}
\end{table}

\begin{table}[t]
    \centering
    \caption{Scoring functions. $d$ denotes a dimension size.}
    \resizebox{0.95\columnwidth}{!}{
    \small
    \begin{tabular}{rll}
        \toprule
         \textbf{Model} & \textbf{Score Function} & \textbf{Parameters} \\
         \midrule
         \multirow{2}{*}{RESCAL} & \multirow{2}{*}{$\mathbf{h}^{\intercal}\mathbf{M}_r\mathbf{t}$} & $\mathbf{h},\mathbf{t}\in \mathbb{R}^d$, \\
         & & $\mathbf{M} \in \mathbb{R}^{d \times d}$ \\
         DistMult & $\mathbf{h}^{\intercal}\mathrm{diag}(\mathbf{r})\mathbf{t}$ & $\mathbf{h},\mathbf{r},\mathbf{t}\in \mathbb{R}^d$ \\
         ComplEx & $\mathrm{Re}(\mathbf{h}^{\intercal}\mathrm{diag}(\mathbf{r})\bar{\mathbf{t}})$ & $\mathbf{h},\mathbf{r},\mathbf{t}\in \mathbb{C}^d$ \\
         TransE & $-||\mathbf{h}+\mathbf{r}-\mathbf{t}||_{p}$ & $\mathbf{h},\mathbf{r},\mathbf{t}\in \mathbb{R}^d$ \\
         RotatE & $-||\mathbf{h} \circ \mathbf{r}-\mathbf{t}||_{p}$ & $\mathbf{h},\mathbf{r},\mathbf{t}\in \mathbb{C}^d$,$|r_i|=1$, \\
         \multirow{3}{*}{HAKE} & \multirow{2}{*}{$-||\mathbf{h} \circ \mathbf{r}-\mathbf{t}||_{p}$} & $\mathbf{h},\mathbf{t}\in \mathbb{R}^d$, $\mathbf{r}\in\mathbb{R}^{d}_{+}$  \\
         &  \multirow{2}{*}{$-\lambda||\sin((\mathbf{h}'+\mathbf{r}'-\mathbf{t}')/2)||_{1}$}  & $\mathbf{h}',\mathbf{r}',\mathbf{t}'\in [0,2\pi)^d$, \\
         & & $\lambda \in \mathbb{R}$  \\
        \bottomrule
    \end{tabular}}
    \label{tab:scoring}
\end{table}

\paragraph{Models}
We chose RESCAL, ComplEx, DistMult, TransE, RotatE, and HAKE~\cite{hake} for our experiments. Table \ref{tab:scoring} shows the scoring functions for each model. From the table, we can see that the value range of RESCAL, ComplEx, and DistMult is $(-\infty,\infty)$, whereas that of TransE and RotatE is $(-\infty,0]$. HAKE's value range is close to $(-\infty,\infty)$ when $\lambda$ is close to $-\infty$, whereas it is close to $(-\infty,0]$ when $\lambda$ is close to $0$. Since $\lambda$ is initialized near 0, a large learning rate is required to make $\lambda$ close to $-\infty$. Thus, we can expect that HAKE pretends like TransE and RotatE in a small learning rate.

\paragraph{Implementation \& Hyperparameters} We used the implementations and hyperparameters reported by \citet{rotate}\footnote{\url{https://github.com/DeepGraphLearning/KnowledgeGraphEmbedding}} for ComplEx, DistMult, TransE, and RotatE and the one reported by \citet{hake}\footnote{\url{https://github.com/zyjcs/KGE-HAE}} for HAKE in our experiments unless otherwise noted in the next subsections. We implemented RESCAL by modifying DistMult by basically inheriting the original hyperparameters. We chose uniform noise as $p_n(y|x)$ for the NS loss in Eqs. (\ref{eq:ns:loss}) and (\ref{eq:ns:loss:kge}). We also applied SANS for model training. We used Adam \cite{adam} as our optimizer.

\begin{figure*}[t!]
\centering
\includegraphics[width=\textwidth]{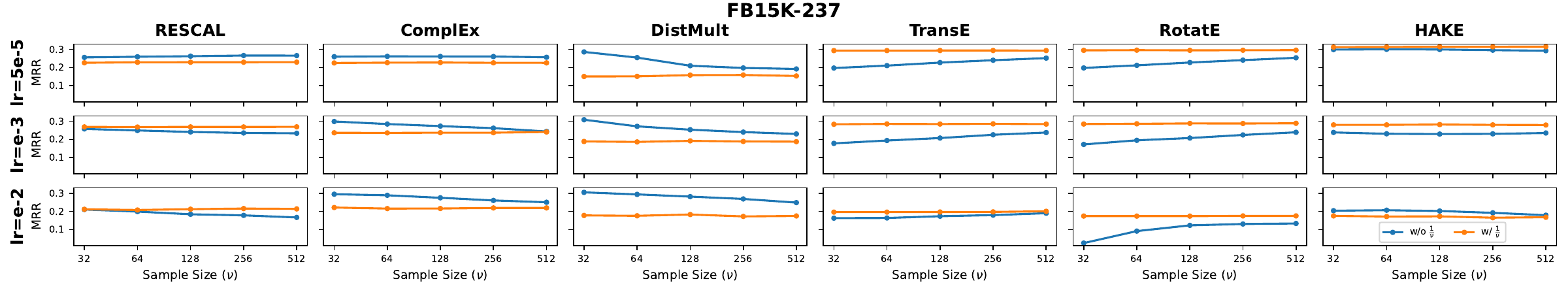}
\includegraphics[width=\textwidth]{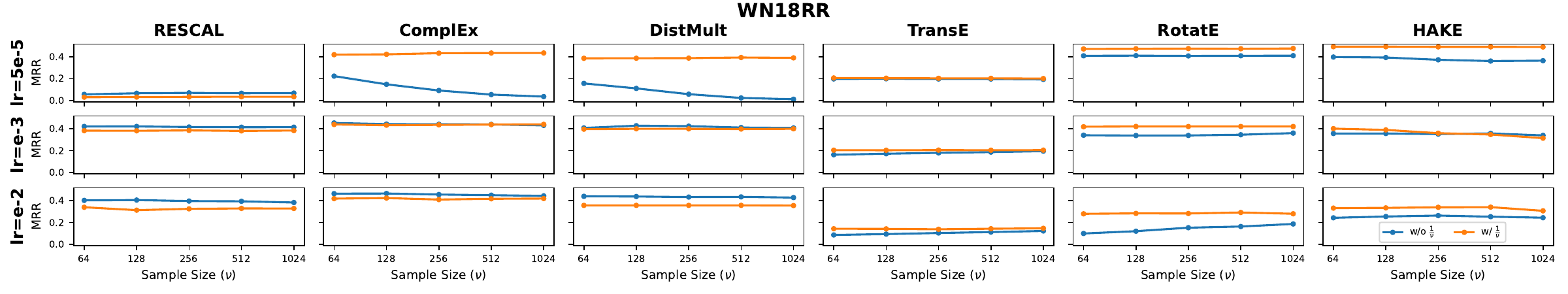}
\caption{MRR scores using zero as margin term $\gamma$ with and without $\frac{1}{\nu}$ normalizing for negative samples in Eq.~(\ref{eq:ns:loss:kge}). \label{fig:sampling:margin:zero}}
\end{figure*}

\subsection{Effects of the Margin Term $\gamma$}
\label{subsec:analysis:margin}

To investigate the performance when changing $\gamma$, we compared the mean reciprocal rank (MRR). We first examined whether the theoretical conclusion that $\gamma$ is necessary to properly learn $p$-norm-based scoring methods as discussed in \S \ref{subsec:margin} is also valid in practice.

\paragraph{Settings} For this investigation, we compared the MRRs of the models introduced in \S \ref{subsec:analysis:settings} on FB15k-237 and WN18RR. We chose the NS loss in Eq.~(\ref{eq:ns:loss:kge}) with uniform noise and the SANS loss to train the models. We chose margin term $\gamma$ from $\{0.0, 3.0, 6.0, 9.0, 9.58\}$ on FB15k-237 and $\{0.0, 2.0, 4.0, 6.0, 10.62\}$ on WN18RR for each initial learning rate in $\{e-2,e-3,5e-5\}$. Note that we decided the set for margin values on the basis of previous studies \cite{rotate,hake} and Eq.~(\ref{eq:margin:condition}). We list the detailed hyperparameters for each model in Appendix \ref{app:analysis:margin:details}.

\paragraph{Results} Figure \ref{fig:margin:zero} shows the results. The results indicate that there was no performance gain for RESCAL, ComplEx, and DistMult by setting non-zero values on $\gamma$, which again confirms no necessity of adjusting $\gamma$ in these scoring methods. We believe the gradient changes by different $\gamma$ caused these performance degradations as shown in Prop.~\ref{prop:grad:gamma}. For TransE and RotatE based on the $p$-norm, there is a significant performance degradation when $\gamma$ decreases, which confirms the necessity of adjusting $\gamma$ with these scoring methods. HAKE also follows the tendencies of TransE and RotatE, especially when the learning rate is small. These tendencies are consistent regardless of the NS loss with uniform noise or SANS loss.

Since the lowest $\gamma$ satisfying Eq.~(\ref{eq:margin:condition}) is $9.58$ for FB15k-237 and $10.62$ for WN18RR, the model cannot reach the optimal solution with $\gamma$ reported in the previous study, $9.0$ for FB15k-237 and $6.0$ for WN18RR. However, the performance gains in MRR when changing $\gamma$ from $9.0$ to $9.58$ on FB15k-237 and from $6.0$ to $10.62$ on WN18RR are slight. The goal of learning is to generalize the model to not fit perfectly to the training data. From this viewpoint, even if the model does not reach the optimal solution, it is not a large problem if it can reach the vicinity of the optimal solution. In fact, the tendencies of losses and MRRs are not perfectly the same.

From the results, we can conclude that in actual training, we should use $\gamma$ to train distance-based scoring methods; however, Eq.~(\ref{eq:margin:condition}) does not have to be strictly followed.

\begin{figure*}[t!]
\centering
\includegraphics[width=\textwidth]{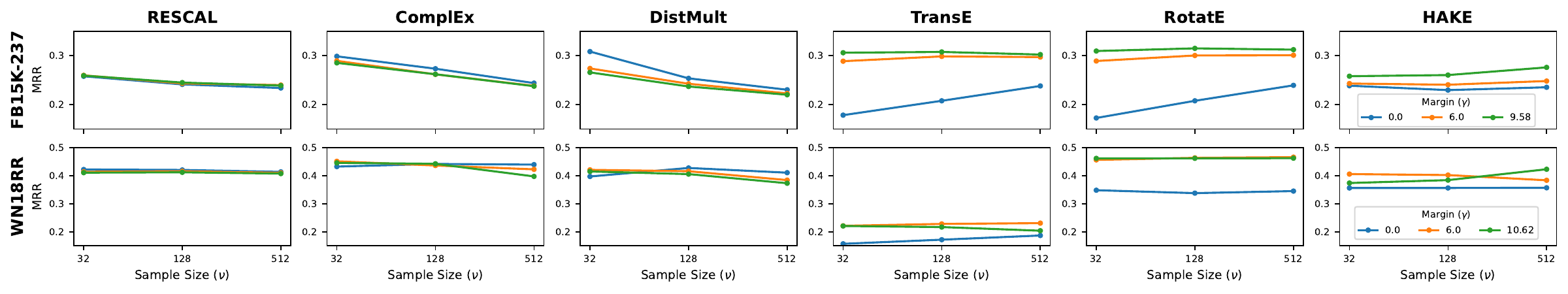}
\caption{ MRR scores for each model of each dataset when we vary negative sample size $\nu$ and margin term $\gamma$.\label{fig:sampling:various}}
\end{figure*}
\begin{figure*}[t!]
\centering
\includegraphics[width=\textwidth]{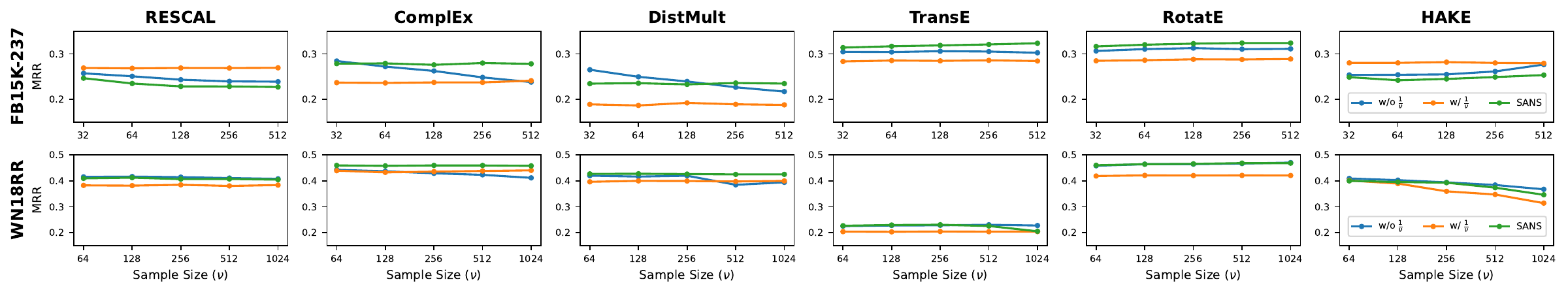}
\caption{MRR scores for each model of each dataset when we vary negative sample size $\nu$. w/o $\frac{1}{\nu}$ and w/ $\frac{1}{\nu}$ represent NS loss in Eqs. (\ref{eq:ns:loss}) and (\ref{eq:ns:loss:kge}), respectively. \label{fig:sans:sampling:various}}
\end{figure*}

\subsection{Effects of the Number of Negative Samples $\nu$}
\label{subsec:analysis:nu}

Next, we investigated the effect of $\nu$ in learning.

\subsubsection{Value ranges of scoring methods}
\label{subsubsec:nu:range}

From the theoretical perspective, $\nu$ helps the distance-based scoring method reach the optimal solution in the original NS loss in Eq. (\ref{eq:ns:loss}) as shown in Prop. \ref{prop:bound2}.
We conducted an experiment to confirm this theoretical finding.

\paragraph{Settings} We compared the MRRs of the models introduced in \S \ref{subsec:analysis:settings} on FB15k-237 and WN18RR. We used the NS loss in Eq. (\ref{eq:ns:loss}) (w/o $\frac{1}{\nu}$). We varied the number of negative samples $\nu$ from $\{32, 64, 128, 256, 512\}$ on FB15k-237 and $\{64, 128, 256, 512, 1024\}$ on WN18RR for each initial learning rate in $\{e-2,e-3,5e-5\}$. We list the detailed hyperparameters for each model in Appendix \ref{app:analysis:nu:details}.

\paragraph{Results} The blue lines of w/o $\frac{1}{\nu}$ in Figure \ref{fig:sampling:margin:zero} show the result. This result indicates that increasing $\nu$ assists in learning distance-based scoring methods TransE and RotatE on FB15k-237. However, the performance improvement on WN18RR by increasing $\nu$ is slight or none. Because WN18RR has a larger number of entities than FB15k-237, we consider that $\nu$ was insufficient for WN18RR. RESCAL, ComplEx, and DistMult have no performance gains by increasing $\nu$. These observations indicate that Prop. \ref{prop:bound2} holds in actual training.

\subsubsection{Gradient changes}

As shown in Prop. \ref{prop:ns}, changing $\nu$ affects the gradient of the NS loss in Eq. (\ref{eq:ns:loss}) but not in Eq. (\ref{eq:ns:loss:kge}).
We investigated the effect of the gradients in the actual training.

\paragraph{Settings} We inherited the setting in \S \ref{subsubsec:nu:range} considering the NS loss in Eq. (\ref{eq:ns:loss:kge}) (w/ $\frac{1}{\nu}$). We set $\gamma$ as $9.0$ on FB15k-237 and $6.0$ on WN18RR.

\paragraph{Results} Figure \ref{fig:sampling:margin:zero} shows the result. By changing $\nu$, we can see the stable performance of the NS loss in Eq.~(\ref{eq:ns:loss:kge}) compared with that in Eq.~(\ref{eq:ns:loss}) for each learning rate. From the result, we can understand that Prop.~\ref{prop:ns} holds in the actual training.

\subsection{Relationship between the Margin Term $\gamma$ and the Number of Negative Samples $\nu$}
\label{subsec:analysis:nu-margin}

Props. \ref{prop:bound1} and \ref{prop:bound2} indicate that margin term $\gamma$ and number of negative samples $\nu$ have a similar role.
Since many factors affect the model performance, we investigated whether this relationship still holds in the actual training.

\paragraph{Settings} In this investigation, we compared the MRRs of the models introduced in \S \ref{subsec:analysis:settings} on FB15k-237 and WN18RR. To consider the effect of both $\nu$ and $\gamma$, we used the NS loss in Eq.~(\ref{eq:ns:loss:kge}) by canceling its normalization term $\frac{1}{\nu}$. We varied $\nu$ from $\{32, 128, 512\}$ and $\gamma$ from $\{0.0, 6.0, 9.58\}$ on FB15k-237 and $\{0.0, 6.0, 10.62\}$ on WN18RR with the commonly used initial learning rate $e-3$ in Adam. We list the detailed hyperparameters for each model in Appendix \ref{app:analysis:nu-margin:details}.

\paragraph{Results} Figure \ref{fig:sampling:various} shows the result. In $p$-norm-based scoring methods TransE and RotatE, we can observe that large $\gamma$ can be a replacement of large $\nu$, whereas the results in the other models do not have such relationships. Therefore, we can consider that $\gamma$ and $\nu$ have a similar role for $p$-norm-based scoring methods in the actual training. Compared with increasing $\gamma$, performance improvements by increasing $\nu$ are slight. This is because $\gamma$ has an exponential effect on the NS loss, while $\nu$ has only a linear effect and is small for reducing the computation time. Thus, using $\gamma$ is more efficient than tuning $\nu$ for training a distance-based scoring method.

\subsection{Effects of Gradient Changes in SANS}
\label{subsec:analysis:sans}

As discussed in Prop.~\ref{prop:sans}, when $\nu$ is sufficiently large to approximate model probabilities, we can assume that $\nu$ does not affect the gradient during training in SANS similar to the NS loss in Eq. (\ref{eq:ns:loss:kge}). On the other hand, as discussed in Prop.~\ref{prop:ns}, $\nu$ affects the gradient during training of the NS loss in Eq.~(\ref{eq:ns:loss}). Since $\nu$ is not the only factor that affects the gradient, it is not clear whether this difference actually affects model's performance. We investigated these aspects through this experiment.

\paragraph{Settings} Similar to the experiment in \S \ref{subsec:analysis:nu}, we compared the MRRs of the models introduced in \S \ref{subsec:analysis:settings} on the basis of the various number of negative samples $\nu$ on FB15k-237 and WN18RR. We used the NS losses in Eqs.~(\ref{eq:ns:loss}) and (\ref{eq:ns:loss:kge}) and the SANS loss to train the models. To ignore the effect discussed in Prop. \ref{prop:bound2}, on the basis of the best setting of the original implementation, we set $\gamma$ as $9.0$ on FB15k-237 and $6.0$ on WN18RR. Furthermore, we also consider margin term $\gamma$ for the NS loss in Eq.~(\ref{eq:ns:loss}), which is in a similar form to the NS loss in Eq.~(\ref{eq:ns:loss:kge}) without normalization term $\frac{1}{\nu}$. We varied $\nu$ from $\{32, 64, 128, 256, 512\}$ on FB15k-237 and $\{64, 128, 256, 512, 1024\}$ on WN18RR with the commonly used initial learning rate $e-3$ in Adam. We list the detailed hyperparameters for each model in Appendix \ref{app:analysis:sans:details}.

\paragraph{Results} Figure \ref{fig:sans:sampling:various} shows the result. We can see the stable MRRs in the NS loss in Eq.~(\ref{eq:ns:loss:kge}), whereas the NS loss in Eq.~(\ref{eq:ns:loss}) shows the large changes of MRRs in RESCAL, ComplEx, and DistMult. In the SANS loss, we can also see the stable MRRs when $\nu$ is 64 or more.
This means that Props. \ref{prop:ns} and \ref{prop:sans} can be satisfied only using small $\nu$. In conclusion, as well as our theoretical interpretation, we found that the effect of $\nu$ on the actual learning is large for the original NS loss in Eq.~(\ref{eq:ns:loss}), whereas it is small for the NS loss in Eq. (\ref{eq:ns:loss:kge}) and SANS loss.

\begin{table}[t!]
\caption{Evaluation results for each subsampling method on FB15k-237. \textit{Sub.} denotes subsampling, \textit{None} denotes model that did not use subsampling, \textit{Base} denotes Eq. (\ref{eq:subsamp:default}), \textit{Freq} denotes Eq. (\ref{eq:subsamp:freq}), and \textit{Uniq} denotes Eq. (\ref{eq:subsamp:uniq}).\label{tab:sub}}
\centering
\small
\resizebox{0.95\columnwidth}{!}{
\begin{tabular}{llcccc}
\toprule
\multicolumn{6}{c}{\textbf{FB15k-237}}                    \\
\midrule
                      \textbf{Model}  & \textbf{Sub}. &        \textbf{MRR}              &  \textbf{Hits@1}     & \textbf{Hits@3}     & \textbf{Hits@10}    \\
\midrule
\multirow{4}{*}{\textbf{RESCAL}} & None & 17.2 & 9.9 & 18.1 & 31.8 \\
 & Base & 22.3 & 13.9 & 24.2 & 39.8 \\
\cmidrule{2-6}
 & Freq & \textbf{26.6} & 17.4 & \textbf{29.4} & \textbf{45.1} \\
 & Uniq & \textbf{26.6} & \textbf{17.6} & 29.3 & 44.9 \\
\midrule
\multirow{4}{*}{\textbf{ComplEx}} & None & 22.4 & 14.0 & 24.2 & 39.5 \\
 & Base & 32.2 & 23.0 & 35.1 & 51.0 \\
\cmidrule{2-6}
 & Freq & \textbf{32.8} & \textbf{23.6} & \textbf{36.1} & 51.2 \\
 & Uniq & 32.7 & 23.5 & 35.8 & \textbf{51.3} \\
\midrule
\multirow{4}{*}{\textbf{DistMult}} & None & 22.2 & 14.0 & 24.0 & 39.4 \\
 & Base & \textbf{30.8} & \textbf{22.1} & \textbf{33.6} & \textbf{48.4} \\
\cmidrule{2-6}
 & Freq & 29.9 & 21.2 & 32.7 & 47.5 \\
 & Uniq & 29.1 & 20.3 & 31.8 & 46.6 \\
\midrule
\multirow{4}{*}{\textbf{TransE}} & None & 33.0 & 22.8 & 37.2 & \textbf{53.0} \\
 & Base & 32.9 & 23.0 & 36.8 & 52.7 \\
\cmidrule{2-6}
 & Freq & \textbf{33.6} & \textbf{24.0} & \textbf{37.3} & 52.9 \\
 & Uniq & 33.5 & 23.9 & \textbf{37.3} & 52.8 \\
\midrule
\multirow{4}{*}{\textbf{RotatE}} & None & 33.1 & 23.1 & 37.1 & 53.1 \\
 & Base & 33.6 & 23.9 & 37.4 & \textbf{53.2} \\
\cmidrule{2-6}
 & Freq & \textbf{34.0} & \textbf{24.5} & \textbf{37.6} & \textbf{53.2} \\
 & Uniq & \textbf{34.0} & \textbf{24.5} & \textbf{37.6} & 53.0 \\
\midrule
\multirow{4}{*}{\textbf{HAKE}} & None & 32.3 & 21.6 & 36.9 & 53.2 \\
 & Base & 34.5 & 24.7 & 38.2 & 54.3 \\
\cmidrule{2-6}
 & Freq & 34.9 & 25.2 & 38.6 & 54.2 \\
 & Uniq & \textbf{35.4} & \textbf{25.8} & \textbf{38.9} & \textbf{54.7} \\
                      \bottomrule
\end{tabular}}
\end{table}
\begin{table}[ht]
%\caption{Evaluation results for each subsampling method in WN18RR. The notations are the same as Table \ref{tab:sub}. \label{tab:sub:wn18rr}}
\caption{Evaluation results for each subsampling method in WN18RR. The notations are the same as those in Table \ref{tab:sub}. \label{tab:sub:wn18rr}}
\centering
\small
\resizebox{0.95\columnwidth}{!}{
\begin{tabular}{llcccc}
\toprule
\multicolumn{6}{c}{\textbf{WN18RR}} \\
\midrule
\textbf{Model}  & \textbf{Sub}. & \textbf{MRR}              &  \textbf{Hits@1}     & \textbf{Hits@3}     & \textbf{Hits@10}    \\
                      \midrule
\multirow{4}{*}{\textbf{RESCAL}} & None & 41.5 & 39.0 & 42.3 & 45.9 \\
 & Base & 43.3 & 40.7 & 44.5 & 48.2 \\
\cmidrule{2-6}
 & Freq & \textbf{44.1} & 41.1 & \textbf{45.6} & \textbf{49.5} \\
 & Uniq & \textbf{44.1} & \textbf{41.4} & 45.5 & \textbf{49.5} \\
\midrule
\multirow{4}{*}{\textbf{ComplEx}} & None & 45.0 & 40.9 & 46.6 & 53.4 \\
 & Base & 47.1 & 42.8 & 48.9 & 55.7 \\
\cmidrule{2-6}
 & Freq & \textbf{47.6} & \textbf{43.3} & 49.3 & \textbf{56.3} \\
 & Uniq & \textbf{47.6} & 43.2 & \textbf{49.5} & \textbf{56.3} \\
\midrule
\multirow{4}{*}{\textbf{DistMult}} & None & 42.4 & 38.3 & 43.6 & 51.0 \\
 & Base & 43.9 & 39.4 & 45.2 & 53.3 \\
\cmidrule{2-6}
 & Freq & \textbf{44.6} & \textbf{40.0} & 45.9 & \textbf{54.4} \\
 & Uniq & \textbf{44.6} & 39.9 & \textbf{46.2} & 54.3 \\
\midrule
\multirow{4}{*}{\textbf{TransE}} & None & 22.6 & 1.8 & 40.1 & 52.3 \\
 & Base & 22.4 & 1.3 & 40.1 & 53.0 \\
\cmidrule{2-6}
 & Freq & 23.0 & 1.9 & 40.7 & \textbf{53.7} \\
 & Uniq & \textbf{23.2} & \textbf{2.2} & \textbf{41.0} & 53.4 \\
\midrule
\multirow{4}{*}{\textbf{RotatE}} & None & 47.3 & 42.6 & 49.1 & 56.7 \\
 & Base & 47.6 & 43.1 & 49.5 & 56.6 \\
\cmidrule{2-6}
 & Freq & 47.8 & 42.9 & \textbf{49.8} & \textbf{57.4} \\
 & Uniq & \textbf{47.9} & \textbf{43.5} & 49.6 & 56.7 \\
\midrule
\multirow{4}{*}{\textbf{HAKE}} & None & 49.1 & 44.5 & 51.1 & 57.8 \\
 & Base & \textbf{49.8} & 45.3 & \textbf{51.6} & 58.2 \\
\cmidrule{2-6}
 & Freq & 49.7 & 45.2 & 51.4 & \textbf{58.5} \\
 & Uniq & \textbf{49.8} & \textbf{45.4} & 51.5 & 58.3 \\
                      \bottomrule
\end{tabular}}
\end{table}
\begin{table}[ht]
%\caption{Evaluation results for each subsampling method in YAGO3-10. The notations are the same as Table \ref{tab:sub}. \label{tab:sub:yago3-10}}
\caption{Evaluation results for each subsampling method in YAGO3-10. The notations are the same as those in Table \ref{tab:sub}.\label{tab:sub:yago3-10}}
\centering
\small
\resizebox{0.95\columnwidth}{!}{
\begin{tabular}{llcccc}
\toprule
\multicolumn{6}{c}{\textbf{YAGO3-10}}   \\
\midrule
\textbf{Model}  & \textbf{Sub}. & \textbf{MRR}              &  \textbf{Hits@1}     & \textbf{Hits@3}     & \textbf{Hits@10}    \\
\midrule
\multirow{4}{*}{\textbf{TransE}} & None & 50.6 & 40.9 & 56.6 & 67.7 \\
 & Base & 51.2 & 41.5 & \textbf{57.6} & \textbf{68.3} \\
\cmidrule{2-6}
 & Freq & 51.3 & 41.9 & 57.2 & 68.1 \\
 & Uniq & \textbf{51.4} & \textbf{42.0} & \textbf{57.6} & 67.9 \\
\midrule
\multirow{4}{*}{\textbf{RotatE}} & None & 50.6 & 41.1 & 56.5 & 67.8 \\
 & Base & 50.8 & 41.8 & 56.5 & 67.6 \\
\cmidrule{2-6}
 & Freq & 51.0 & 41.9 & 56.5 & 67.8 \\
 & Uniq & \textbf{51.5} & \textbf{42.5} & \textbf{56.8} & \textbf{68.3} \\
\midrule
\multirow{4}{*}{\textbf{HAKE}} & None & 53.4 & 44.9 & 58.7 & 68.4 \\
 & Base & 54.3 & 46.1 & 59.5 & 69.2 \\
\cmidrule{2-6}
 & Freq & 54.0 & 45.5 & 59.4 & 69.1 \\
 & Uniq & \textbf{55.0} & \textbf{46.6} & \textbf{60.1} & \textbf{69.8} \\
                      \bottomrule
\end{tabular}}
\end{table}

\subsection{Effects of Subsampling Methods}

We evaluated the subsampling methods discussed in \S \ref{subsec:subsamp}. We compared the currently used subsampling method derived from Eq.~(\ref{eq:subsamp:default}): \textit{Base} and the proposed methods derived from Eq.~(\ref{eq:subsamp:freq}): \textit{Freq} and (\ref{eq:subsamp:uniq}): \textit{Uniq}. We also included models trained without subsampling in this comparison.

\paragraph{Settings} To train the models, we used the best configuration reported by \citet{rotate} for ComplEx, DistMult, TransE, and RotatE and \citet{hake} for HAKE since they used the subsampling method in Eq. (\ref{eq:subsamp:default}). In RESCAL, which was not included in the original implementation, we inherited the setting of DistMult by reducing its dimension size since RESCAL uses weight matrices instead of vectors for representing each relation. Note that on YAGO3-10, they only reported the configuration of TransE, RotatE, and HAKE. Therefore, we compared only TransE, RotatE, and HAKE on YAGO3-10. Subsampling is a smoothing method, and \citet{unify} showed that the SANS loss also has a smoothing effect. Therefore, to confirm whether the subsampling method is effective in actual training, it is necessary to apply it to the SANS loss. Since the subsampling method's implementation in the previous studies also adjusts the hyperparameters on the basis of the SANS loss, we used the SANS loss in this experiment. We used MRR, Hits@1, Hits@3, and Hits@10 as the evaluation metrics.

\paragraph{Results} Tables \ref{tab:sub}, \ref{tab:sub:wn18rr}, and \ref{tab:sub:yago3-10} show the results. We can see that for MRR, Hits@1, and Hits@3, subsampling was effective for all datasets and all models used in this experiment. Also, for Hits@10, the score was higher when we used subsampling for all models except RotatE. For RotatE, Freq and Uniq achieved a comparable or higher performance in Hits@10 as those without subsampling. This observation suggest that the use of subsampling basically improves KGC performance.

When comparing the results of the subsampling methods, \textit{Freq} and \textit{Uniq} performed as well as or better than \textit{Base} in many cases. This result shows the validity of deriving \textit{Freq} and \textit{Uniq} from the theoretical perspective. However, the effect of the subsampling method varies depending on the combined dataset and model. Considering the theoretical perspective, since we derived \textit{Freq} and \textit{Uniq} by assuming the true distribution behind the dataset, the fact that the results change depending on the model selected is considered contrary to the original intention. This is because several models have already taken into account the training data bias to be corrected by the subsampling method. Therefore, when using a subsampling method, it is important to check the performance of the combination of the model, subsampling method, and dataset by using a development dataset.

\section{Related Work}

\citet{ns} originally proposed the NS loss to approximate the softmax cross-entropy loss function for reducing the computational cost in word-representation learning \citep{w2v}. Initially, due to its similarity to noise contrastive estimation (NCE) \citep{nce}, \citet{nce-ns} conducted a theoretical analysis to determine the differences between the NS loss and NCE loss, but the distinct properties of the NS loss have not been clarified. In particular, the work of \citet{w2vmf} attracted attention because it showed that the NS loss in word2vec is equivalent to decomposing the pointwise mutual information (PMI) matrix by using matrix factorization. However, this equivalence is only valid when the unigram distribution is used as the noise distribution; thus, this finding is restricted to word representation.

The NS loss has been widely used in KGE \citep{complex}. \citet{rotate} proposed SANS for efficiently training KGE models. Since various scoring methods and loss functions are used in combination for KGE, the properties of the loss functions are not clear. To solve this problem, \citet{olddog} validated these combinations through large-scale experiments and empirically derived which combinations are best. \citet{benchmarking} also investigated the best configuration of the current KGE models. \citet{zhang-etal-2022-efficient} proposed fast hyperparameter tuning in KGE.

However, these empirical approaches still require much computational time for hyperparameter tuning. \citet{unify} clarified the properties of the softmax cross-entropy (SCE) loss function and NS loss function by theoretically identifying the differences between them. While their discussion covers a wide range of applications from word2vec to KGE, they did not include the NS loss specifically used for KGE in their analysis.

Our work is novel in that we analyzed the NS loss theoretically considering $\gamma$ and the normalization of $\nu$ used in KGE in previous studies. These points have not been discussed in those studies. In addition, our study included the SANS loss, specifically proposed for KGE learning, in the discussion. Thus, our discussion is useful for training scoring methods using SANS. We also proposed subsampling methods specialized for KGE that can actually improve KGC performance.

\section{Conclusion}

We conducted a theoretical analysis for the NS loss used in KGE learning and derived theoretical facts. On the basis of these facts, we also proposed subsampling methods suitable for the NS loss function of KGE learning. To verify the validity of these theoretical facts and the performance of the proposed subsampling methods, we conducted experiments on the FB15k-237, WN18RR, and YAGO3-10 datasets. The experimental results indicate that the theoretical facts we derived are also observed in the real-world datasets.

\section*{Acknowledgements}
The authors thank anonymous reviewers for insightful comments and suggestions.
This work was partially supported by JSPS Kakenhi
Grant Nos. 21H03491, 21K17801.

\if0
\textbf{Do not} include acknowledgements in the initial version of
the paper submitted for blind review.

If a paper is accepted, the final camera-ready version can (and
probably should) include acknowledgements. In this case, please
place such acknowledgements in an unnumbered section at the
end of the paper. Typically, this will include thanks to reviewers
who gave useful comments, to colleagues who contributed to the ideas,
and to funding agencies and corporate sponsors that provided financial
support.
\fi

\nocite{langley00}

\bibliography{example_paper}

\begin{thebibliography}{21}
\providecommand{\natexlab}[1]{#1}
\providecommand{\url}[1]{\texttt{#1}}
\expandafter\ifx\csname urlstyle\endcsname\relax
  \providecommand{\doi}[1]{doi: #1}\else
  \providecommand{\doi}{doi: \begingroup \urlstyle{rm}\Url}\fi

\bibitem[Ahrabian et~al.(2020)Ahrabian, Feizi, Salehi, Hamilton, and
  Bose]{ahrabian-etal-2020-structure}
Ahrabian, K., Feizi, A., Salehi, Y., Hamilton, W.~L., and Bose, A.~J.
\newblock Structure aware negative sampling in knowledge graphs.
\newblock In \emph{Proceedings of the 2020 Conference on Empirical Methods in
  Natural Language Processing (EMNLP)}, pp.\  6093--6101, Online, November
  2020. Association for Computational Linguistics.
\newblock \doi{10.18653/v1/2020.emnlp-main.492}.
\newblock URL \url{https://aclanthology.org/2020.emnlp-main.492}.

\bibitem[Ali et~al.(2021)Ali, Berrendorf, Hoyt, Vermue, Galkin, Sharifzadeh,
  Fischer, Tresp, and Lehmann]{benchmarking}
Ali, M., Berrendorf, M., Hoyt, C.~T., Vermue, L., Galkin, M., Sharifzadeh, S.,
  Fischer, A., Tresp, V., and Lehmann, J.
\newblock Bringing light into the dark: A large-scale evaluation of knowledge
  graph embedding models under a unified framework.
\newblock \emph{IEEE Transactions on Pattern Analysis and Machine
  Intelligence}, pp.\  1--1, 2021.
\newblock \doi{10.1109/TPAMI.2021.3124805}.

\bibitem[Bordes et~al.(2011)Bordes, Weston, Collobert, and Bengio]{rescal}
Bordes, A., Weston, J., Collobert, R., and Bengio, Y.
\newblock Learning structured embeddings of knowledge bases.
\newblock In \emph{Proceedings of the Twenty-Fifth AAAI Conference on
  Artificial Intelligence}, AAAI'11, pp.\  301–306. AAAI Press, 2011.

\bibitem[Bordes et~al.(2013)Bordes, Usunier, Garc{\'{\i}}a{-}Dur{\'{a}}n,
  Weston, and Yakhnenko]{transe}
Bordes, A., Usunier, N., Garc{\'{\i}}a{-}Dur{\'{a}}n, A., Weston, J., and
  Yakhnenko, O.
\newblock Translating embeddings for modeling multi-relational data.
\newblock In \emph{Advances in Neural Information Processing Systems 26: 27th
  Annual Conference on Neural Information Processing Systems 2013}, pp.\
  2787--2795, 2013.
\newblock URL
  \url{https://proceedings.neurips.cc/paper/2013/hash/1cecc7a77928ca8133fa24680a88d2f9-Abstract.html}.

\bibitem[Dettmers et~al.(2018)Dettmers, Minervini, Stenetorp, and
  Riedel]{conve}
Dettmers, T., Minervini, P., Stenetorp, P., and Riedel, S.
\newblock Convolutional 2d knowledge graph embeddings.
\newblock In \emph{Proceedings of the Thirty-Second {AAAI} Conference on
  Artificial Intelligence, (AAAI-18)}, pp.\  1811--1818, 2018.
\newblock URL
  \url{https://www.aaai.org/ocs/index.php/AAAI/AAAI18/paper/view/17366}.

\bibitem[Dyer(2014)]{nce-ns}
Dyer, C.
\newblock Notes on noise contrastive estimation and negative sampling.
\newblock \emph{arXiv preprint arXiv:1410.8251}, 2014.

\bibitem[Gutmann \& Hyv{\"a}rinen(2010)Gutmann and Hyv{\"a}rinen]{nce}
Gutmann, M. and Hyv{\"a}rinen, A.
\newblock Noise-contrastive estimation: A new estimation principle for
  unnormalized statistical models.
\newblock In \emph{Proceedings of the Thirteenth International Conference on
  Artificial Intelligence and Statistics}, pp.\  297--304, 2010.

\bibitem[Kamigaito \& Hayashi(2021)Kamigaito and Hayashi]{unify}
Kamigaito, H. and Hayashi, K.
\newblock Unified interpretation of softmax cross-entropy and negative
  sampling: With case study for knowledge graph embedding.
\newblock In \emph{Proceedings of the 59th Annual Meeting of the Association
  for Computational Linguistics and the 11th International Joint Conference on
  Natural Language Processing, {ACL/IJCNLP} 2021}, pp.\  5517--5531, 2021.
\newblock \doi{10.18653/v1/2021.acl-long.429}.
\newblock URL \url{https://doi.org/10.18653/v1/2021.acl-long.429}.

\bibitem[Katz(1987)]{katz1987estimation}
Katz, S.
\newblock Estimation of probabilities from sparse data for the language model
  component of a speech recognizer.
\newblock \emph{IEEE transactions on acoustics, speech, and signal processing},
  35\penalty0 (3):\penalty0 400--401, 1987.

\bibitem[Kingma \& Ba(2015)Kingma and Ba]{adam}
Kingma, D.~P. and Ba, J.
\newblock Adam: {A} method for stochastic optimization.
\newblock In Bengio, Y. and LeCun, Y. (eds.), \emph{3rd International
  Conference on Learning Representations, {ICLR} 2015, San Diego, CA, USA, May
  7-9, 2015, Conference Track Proceedings}, 2015.
\newblock URL \url{http://arxiv.org/abs/1412.6980}.

\bibitem[Kloek \& Van~Dijk(1978)Kloek and Van~Dijk]{kloek1978bayesian}
Kloek, T. and Van~Dijk, H.~K.
\newblock Bayesian estimates of equation system parameters: an application of
  integration by monte carlo.
\newblock \emph{Econometrica: Journal of the Econometric Society}, pp.\  1--19,
  1978.

\bibitem[Levy \& Goldberg(2014)Levy and Goldberg]{w2vmf}
Levy, O. and Goldberg, Y.
\newblock Neural word embedding as implicit matrix factorization.
\newblock In Ghahramani, Z., Welling, M., Cortes, C., Lawrence, N., and
  Weinberger, K.~Q. (eds.), \emph{Advances in Neural Information Processing
  Systems}, volume~27. Curran Associates, Inc., 2014.
\newblock URL
  \url{https://proceedings.neurips.cc/paper/2014/file/feab05aa91085b7a8012516bc3533958-Paper.pdf}.

\bibitem[Mikolov et~al.(2013{\natexlab{a}})Mikolov, Chen, Corrado, and
  Dean]{w2v}
Mikolov, T., Chen, K., Corrado, G., and Dean, J.
\newblock Efficient estimation of word representations in vector space.
\newblock In \emph{Proceedings of the 1st International Conference on Learning
  Representations, {ICLR} 2013}, 2013{\natexlab{a}}.
\newblock URL \url{http://arxiv.org/abs/1301.3781}.

\bibitem[Mikolov et~al.(2013{\natexlab{b}})Mikolov, Sutskever, Chen, Corrado,
  and Dean]{ns}
Mikolov, T., Sutskever, I., Chen, K., Corrado, G., and Dean, J.
\newblock Distributed representations of words and phrases and their
  compositionality.
\newblock \emph{CoRR}, abs/1310.4546, 2013{\natexlab{b}}.
\newblock URL \url{http://arxiv.org/abs/1310.4546}.

\bibitem[Ruffinelli et~al.(2020)Ruffinelli, Broscheit, and Gemulla]{olddog}
Ruffinelli, D., Broscheit, S., and Gemulla, R.
\newblock You {CAN} teach an old dog new tricks! on training knowledge graph
  embeddings.
\newblock In \emph{Proceedings of the 8th International Conference on Learning
  Representations, {ICLR} 2020}, 2020.
\newblock URL \url{https://openreview.net/forum?id=BkxSmlBFvr}.

\bibitem[Sun et~al.(2019)Sun, Deng, Nie, and Tang]{rotate}
Sun, Z., Deng, Z., Nie, J., and Tang, J.
\newblock Rotate: Knowledge graph embedding by relational rotation in complex
  space.
\newblock In \emph{Proceedings of the 7th International Conference on Learning
  Representations, {ICLR} 2019}, 2019.
\newblock URL \url{https://openreview.net/forum?id=HkgEQnRqYQ}.

\bibitem[Toutanova \& Chen(2015)Toutanova and Chen]{fb15k-237}
Toutanova, K. and Chen, D.
\newblock Observed versus latent features for knowledge base and text
  inference.
\newblock In \emph{Proceedings of the 3rd Workshop on Continuous Vector Space
  Models and their Compositionality}, pp.\  57--66, Beijing, China, July 2015.
  Association for Computational Linguistics.
\newblock \doi{10.18653/v1/W15-4007}.
\newblock URL \url{https://www.aclweb.org/anthology/W15-4007}.

\bibitem[Trouillon et~al.(2016)Trouillon, Welbl, Riedel, Gaussier, and
  Bouchard]{complex}
Trouillon, T., Welbl, J., Riedel, S., Gaussier, {\'{E}}., and Bouchard, G.
\newblock Complex embeddings for simple link prediction.
\newblock In \emph{Proceedings of the 33nd International Conference on Machine
  Learning, {ICML} 2016}, volume~48 of \emph{{JMLR} Workshop and Conference
  Proceedings}, pp.\  2071--2080. JMLR.org, 2016.
\newblock URL \url{http://proceedings.mlr.press/v48/trouillon16.html}.

\bibitem[Yang et~al.(2015)Yang, Yih, He, Gao, and Deng]{distmult}
Yang, B., Yih, W., He, X., Gao, J., and Deng, L.
\newblock Embedding entities and relations for learning and inference in
  knowledge bases.
\newblock In \emph{Proceddings of the 3rd International Conference on Learning
  Representations, {ICLR} 2015}, 2015.
\newblock URL \url{http://arxiv.org/abs/1412.6575}.

\bibitem[Zhang et~al.(2022)Zhang, Zhou, Yao, and Li]{zhang-etal-2022-efficient}
Zhang, Y., Zhou, Z., Yao, Q., and Li, Y.
\newblock Efficient hyper-parameter search for knowledge graph embedding.
\newblock In \emph{Proceedings of the 60th Annual Meeting of the Association
  for Computational Linguistics (Volume 1: Long Papers)}, pp.\  2715--2735,
  Dublin, Ireland, May 2022. Association for Computational Linguistics.
\newblock \doi{10.18653/v1/2022.acl-long.194}.
\newblock URL \url{https://aclanthology.org/2022.acl-long.194}.

\bibitem[Zhang et~al.(2020)Zhang, Cai, Zhang, and Wang]{hake}
Zhang, Z., Cai, J., Zhang, Y., and Wang, J.
\newblock Learning hierarchy-aware knowledge graph embeddings for link
  prediction.
\newblock In \emph{Proceedings of the Thirty-Fourth {AAAI} Conference on
  Artificial Intelligence, (AAAI20)}, pp.\  3065--3072, 2020.

\end{thebibliography}
\bibliographystyle{icml2022}

\newpage
\appendix
\onecolumn
\section{The Proofs of the Propositions}

\subsection{The Proof for Prop.~\ref{prop:obj}}
\label{app:subsec:obj}
The previous studies \cite{w2vmf,unify} show that the minimization for Eq.~(\ref{eq:ns:loss}) induces the following equation:
\begin{equation}
    \exp(s_{\theta}(x,y))=\frac{p_d(y|x)}{\nu p_n(y|x)}.
    \label{eq:opt:w2v}
\end{equation}
Now we consider the minimization for the NS loss in Eq.~(\ref{eq:ns:loss:kge}). The normalization to $\nu$ in the NS loss in Eq.~(\ref{eq:ns:loss:kge}) eliminates $\nu$ in Eq.~(\ref{eq:opt:w2v}). Furthermore, due to the existence of the margin term $\gamma$, the minimization for the NS loss in Eq.~(\ref{eq:ns:loss:kge}) induces the following equation:
\begin{equation}
    \exp(s_{\theta}(x,y))=\frac{p_d(y|x)}{\exp(\gamma) p_n(y|x)}.
    \label{eq:opt:kge}
\end{equation}
Here, substituting Eq.~(\ref{eq:opt:w2v}) into Eq.~(\ref{eq:softmax}) and substituting Eq.~(\ref{eq:opt:kge}) into Eq.~(\ref{eq:softmax}) both result in the following formula:
\begin{equation}
    \frac{p_d(y|x)/p_n(y|x)}{\sum_{y'\in Y}{(p_d(y'|x)/p_n(y'|x))}}.
    \label{eq:obj:dist}    
\end{equation}
Therefore, from Eq.~(\ref{eq:obj:dist}), we can understand that Prop.~\ref{prop:obj} holds.

\subsection{The Proof for Props. \ref{prop:bound1} and \ref{prop:bound2}}
\label{app:subsec:bound}
When $s_{\theta}(x,y)\leqq 0$, then $\exp(s_{\theta}(x,y))\leqq 1$ holds. If the right-hand side of Eq.~(\ref{eq:opt:w2v}) do not have the same value range as $\exp(s_{\theta}(x,y))$, then the NS loss in Eq.~(\ref{eq:ns:loss}) can have an unreachable point. Therefore, we can derive the condition that the NS loss in Eq.~(\ref{eq:ns:loss}) does not have an unreachable point as follows:
\begin{align}
    & \frac{p_d(y|x)}{\nu p_n(y|x)}\leqq 1 \nonumber\\
    \Leftrightarrow & p_d(y|x) \leqq \nu p_n(y|x).
    \label{eq:cond1}
\end{align}
We can similarly induce the condition that the NS loss in Eq.~(\ref{eq:ns:loss:kge}) has no unreachable point as follows:
\begin{align}
    & \frac{p_d(y|x)}{\exp(\gamma) p_n(y|x)}\leqq 1 \nonumber\\
    \Leftrightarrow & p_d(y|x) \leqq \exp(\gamma) p_n(y|x).
    \label{eq:cond2}
\end{align}

Here, from Eq.~(\ref{eq:cond1}) and (\ref{eq:cond2}), we can understand that Prop. \ref{prop:bound1} and \ref{prop:bound2} hold.

\subsection{The Proof for Prop. \ref{prop:grad:gamma}}
\label{app:subsec:grad:gamma}

From the derivative of a composite function, it is obvious that the result of differentiating the NS loss in Eq.~(\ref{eq:ns:loss:kge}) contains $\gamma$. Thus, we can understand that Prop. \ref{prop:grad:gamma} holds.

\subsection{The Proof for Prop. \ref{prop:ns}}
\label{app:subsec:ns}

In the NS loss of Eq.~(\ref{eq:ns:loss}), we can consider the following approximation by utilizing the Monte Carlo method:
\begin{equation}
    \sum_{y}p_n(y|x)\log(\sigma(-s_{\theta}(x,y)))\approx\frac{1}{\nu}\sum_{i=1,y_{i}\sim p_n(y_{i}|x)}^{\nu}\log(\sigma(-s_{\theta}(x,y_i))).
    \label{eq:mc:ex}
\end{equation}
Based on Eq.~(\ref{eq:mc:ex}), we can reformulate Eq.~(\ref{eq:ns:loss}) as follows:
\begin{align}
& -\frac{1}{|D|}\sum_{(x,y) \in D} \Bigl[\log(\sigma(s_{\theta}(x,y))) + \sum_{y_{i}\sim p_n(y_{i}|x)}^{\nu}\log(\sigma(-s_{\theta}(x,y_i)))\Bigr]\nonumber\\
\approx & -\frac{1}{|D|}\sum_{(x,y) \in D} \Bigl[\log(\sigma(s_{\theta}(x,y))) + \nu\sum_{y_{i}}p_n(y_i|x)\log(\sigma(-s_{\theta}(x,y_i)))\Bigr].
\label{eq:loss:ref}
\end{align}
In the NS loss of Eq.~(\ref{eq:ns:loss:kge}), we can similarly consider the following approximation by utilizing the Monte Carlo method:
\begin{equation}
    \sum_{y}p_n(y|x)\log(\sigma(-s_{\theta}(x,y)-\gamma)) \approx \frac{1}{\nu}\sum_{i=1,y_{i}\sim p_n(y_{i}|x)}^{\nu}\log(\sigma(-s_{\theta}(x,y_i)-\gamma)).
    \label{eq:mc:ex:gamma}
\end{equation}
Based on Eq.~(\ref{eq:mc:ex:gamma}), we can reformulate Eq.~(\ref{eq:ns:loss:kge}) as follows:
\begin{align}
    & -\frac{1}{|D|}\sum_{(x,y) \in D} \Bigl[\log(\sigma(s_{\theta}(x,y)+\gamma)) + \frac{1}{\nu}\sum_{y_{i}\sim p_n(y_{i}|x)}^{\nu}\log(\sigma(-s_{\theta}(x,y_i)-\gamma))\Bigr]\nonumber\\
    \approx & -\frac{1}{|D|}\sum_{(x,y) \in D} \Bigl[\log(\sigma(s_{\theta}(x,y)+\gamma)) + \sum_{y_{i}}p_n(y_i|x)\log(\sigma(-s_{\theta}(x,y_i)-\gamma))\Bigr].
    \label{eq:loss:kge:ref}
\end{align}
Since the reformulation of Eq.~(\ref{eq:loss:ref}) and (\ref{eq:loss:kge:ref}) is valid, when $\nu$ is enough large, we can understand that Prop. \ref{prop:ns} holds.

\subsection{The Proof for Prop. \ref{prop:sans}}
\label{app:subsec:sans}
Self-adversarial negative sampling (SANS) is defined as follows:
\begin{equation}
-\frac{1}{|D|}\sum_{(x,y) \in D} \Bigl[\log(\sigma(s_{\theta}(x,y)+\gamma)) + \sum_{i=1,y_{i}\sim uniform}^{\nu}\tilde{p}_{\theta}(y_i|x)\log(\sigma(-s_{\theta}(x,y_i)-\gamma))\Bigr],
\label{eq:sans}
\end{equation}
where $\tilde{p}_{\theta}(y_i|x)$ is an approximated $p_{\theta}(y_{i}|x)$ with sampled $y_j$, defined as follows:
\begin{equation}
    \tilde{p}_{\theta}(y_i|x)=\frac{\exp{(s_{\theta}(x,y_i))}}{\sum_{j=1}^{\nu}{\exp{(s_{\theta}(x,y_j))}}}.
\end{equation}
Here, we can reformulate Eq.~(\ref{eq:sans}) as follows:
\begin{align}
&-\frac{1}{|D|}\sum_{(x,y) \in D}\Bigl[\log(\sigma(s_{\theta}(x,y)+\gamma)) + \sum_{i=1,y_{i}\sim uniform}^{\nu}\tilde{p}_{\theta}(y_i|x)\log(\sigma(-s_{\theta}(x,y_i)-\gamma))\Bigr]\nonumber\\
=&-\frac{1}{|D|}\sum_{(x,y) \in D}\Bigl[\log(\sigma(s_{\theta}(x,y)+\gamma)) + \frac{|Y|}{\nu}\sum_{i=1,y_{i}\sim uniform}^{\nu}\frac{\exp(s_{\theta}(x,y_{i}))}{\frac{|Y|}{\nu}\sum_{j=1}^{\nu}\exp(s_{\theta}(x,y_{j}))}\log(\sigma(-s_{\theta}(x,y_i)-\gamma))\Bigr].
\label{eq:sans:ref1}
\end{align}
In Eq.~(\ref{eq:sans:ref1}), we can consider $\frac{|Y|}{\nu}\sum_{j=1}^{\nu}\exp(s_{\theta}(x,y_{j}))$ as a Monte Carlo expectation to approximate $\sum_{j=1}^{|Y|}\exp(s_{\theta}(x,y_{j}))$ in $p_{\theta}(y_i|x)$. Thus, we can reformulate Eq.~(\ref{eq:sans:ref1}) as follows: 
\begin{align}
(\ref{eq:sans:ref1})\approx&-\frac{1}{|D|}\sum_{(x,y) \in D}\Bigl[\log(\sigma(s_{\theta}(x,y)+\gamma))+\frac{|Y|}{\nu}\sum_{i=1,y_{i}\sim uniform}^{\nu}p_{\theta}(y_i|x)\log(\sigma(-s_{\theta}(x,y_i)-\gamma))\Bigr]\nonumber\\
=&-\frac{1}{|D|}\sum_{(x,y)\in D}\Bigl[\log(\sigma(s_{\theta}(x,y)+\gamma)) + \frac{1}{\nu}\sum_{i=1,y_{i}\sim uniform}^{\nu}\frac{p_{\theta}(y_i|x)}{\frac{1}{|Y|}}\log(\sigma(-s_{\theta}(x,y_i)-\gamma))\Bigr].
\label{eq:sans:ref2}
\end{align}
By using importance sampling \citep{kloek1978bayesian}, we can consider the following approximation:
\begin{align}
    &\frac{1}{\nu}\sum_{i=1,y_{i}\sim uniform}^{\nu}\frac{p_{\theta}(y_i|x)}{\frac{1}{|Y|}}\log(\sigma(-s_{\theta}(x,y_i)-\gamma))\nonumber\\
    \approx &\sum_{y_{i}}\frac{1}{|Y|}\frac{p_{\theta}(y_i|x)}{\frac{1}{|Y|}}\log(\sigma(-s_{\theta}(x,y_i)-\gamma))\nonumber\\
    = &\sum_{y_{i}}p_{\theta}(y_i|x)\log(\sigma(-s_{\theta}(x,y_i)-\gamma))
    \label{eq:approx:is}
\end{align}

Based on Eq.~(\ref{eq:approx:is}), we can further reformulate Eq.~(\ref{eq:sans:ref2}) as follows: 
\begin{align}
(\ref{eq:sans:ref2})\approx&-\frac{1}{|D|}\sum_{(x,y)\in D}\Bigl[\log(\sigma(s_{\theta}(x,y)+\gamma)) + \sum_{y_{i}}p_{\theta}(y_i|x)\log(\sigma(-s_{\theta}(x,y_i)-\gamma))\Bigr].
\label{eq:sans:derive}
\end{align}
Since Eq.~(\ref{eq:sans:derive}) is the same when $p_{n}(y|x)=p_{\theta}(y|x)$ in Eq.~(\ref{eq:loss:kge:ref}), we can understand that Prop. \ref{prop:sans} holds.

\section{The Detailed Derivation of Eq.~(\ref{eq:subsamp})}
\label{app:derivation}
We can reformulate the NS loss in Eq.~(\ref{eq:ns:loss:kge}) as follows:
\begin{align}
(\ref{eq:ns:loss:kge})=& -\frac{1}{|D|}\sum_{(x,y) \in D} \Bigl[\log(\sigma(s_{\theta}(x,y)+\gamma)) + \frac{1}{\nu}\sum_{y_{i}\sim p_n(y_{i}|x)}^{\nu}\log(\sigma(-s_{\theta}(x,y_i)-\gamma))\Bigr]\nonumber\\
= & -\frac{1}{|D|}\sum_{(x,y) \in D} \log(\sigma(s_{\theta}(x,y)+\gamma)) -\frac{1}{|D|}\sum_{(x,y) \in D}\frac{1}{\nu}\sum_{y_{i}\sim p_n(y_{i}|x)}^{\nu}\log(\sigma(-s_{\theta}(x,y_i)-\gamma)).
\label{eq:app:sub:intro}
\end{align}
Here, we can consider the following approximation based on the Monte Carlo method:
\begin{equation}
    \frac{1}{\nu}\sum_{y_{i}\sim p_n(y_{i}|x)}^{\nu}\log(\sigma(-s_{\theta}(x,y_i)-\gamma)) \approx \sum_{y}p_n(y|x)\log(\sigma(-s_{\theta}(x,y)-\gamma)).
    \label{eq:app:sub:intro:approx}
\end{equation}
Using Eq.~(\ref{eq:app:sub:intro:approx}), we can reformulate Eq.~(\ref{eq:app:sub:intro}) as follows: 
\begin{equation}
(\ref{eq:app:sub:intro}) \approx -\frac{1}{|D|}\sum_{(x,y) \in D} \log(\sigma(s_{\theta}(x,y)+\gamma)) -\frac{1}{|D|}\sum_{(x,y) \in D}\sum_{y'}p_n(y'|x)\log(\sigma(-s_{\theta}(x,y')-\gamma)).
\label{eq:app:sub:intro2}
\end{equation}
Similr to Eq.~(\ref{eq:app:sub:intro:approx}), we can consider the following approximation by the the Monte Carlo method:
\begin{align}
    -&\frac{1}{|D|}\sum_{(x,y) \in D} \log(\sigma(s_{\theta}(x,y)+\gamma))\approx -\sum_{x,y}\log(\sigma(s_{\theta}(x,y)+\gamma))p_d(x,y), \nonumber\\
    -&\frac{1}{|D|}\sum_{(x,y) \in D}\sum_{y'}p_n(y'|x)\log(\sigma(-s_{\theta}(x,y')-\gamma))\approx -\sum_{x}\sum_{y'}p_n(y'|x)\log(\sigma(-s_{\theta}(x,y')-\gamma))p_d(x).
    \label{eq:app:sub:approx2}
\end{align}
Using Eq.~(\ref{eq:app:sub:approx2}), we can reformulate Eq.~(\ref{eq:app:sub:intro2}) as follows:
\begin{align}
(\ref{eq:app:sub:intro2}) \approx & -\sum_{x,y}\log(\sigma(s_{\theta}(x,y)+\gamma))p_d(x,y) -\sum_{x}\sum_{y'}p_n(y'|x)\log(\sigma(-s_{\theta}(x,y')-\gamma))p_d(x)\nonumber\\
= & -\sum_{x,y}\log(\sigma(s_{\theta}(x,y)+\gamma))p_d(x,y) -\sum_{x,y}p_n(y|x)\log(\sigma(-s_{\theta}(x,y)-\gamma))p_d(x)\nonumber\\
= & -\sum_{x,y} \Bigl[\log(\sigma(s_{\theta}(x,y)+\gamma))p_{d}(x,y) +p_n(y|x)\log(\sigma(-s_{\theta}(x,y)-\gamma))p_{d}(x)\Bigr].
\label{eq:app:sub:intro:end}
\end{align}
Next, we consider replacements of $p_{d}(x,y)$ with $p'_{d}(x,y)$ and $p_{d}(x)$ with $p'_{d}(x)$.
By assuming two functions, $A(x,y)$ and $B(x)$, that convert $p_{d}(x,y)$ into $p'_{d}(x,y)$ and $p_{d}(x)$ into $p'_{d}(x)$, we further reformulate Eq.~(\ref{eq:app:sub:intro:end}) as follows:
\begin{align}
& -\sum_{x,y} \Bigl[\log(\sigma(s_{\theta}(x,y)+\gamma))p'_{d}(x,y) +p_n(y|x)\log(\sigma(-s_{\theta}(x,y)-\gamma))p'_{d}(x)\Bigr]\nonumber\\
=&-\sum_{x,y} \Bigl[\log(\sigma(s_{\theta}(x,y)+\gamma))A(x,y)p_{d}(x,y)+p_n(y|x)\log(\sigma(-s_{\theta}(x,y)-\gamma))B(x)p_{d}(x)\Bigr].
\label{eq:app:sub:replaced}
\end{align}
Based on the similar derivation from Eq.~(\ref{eq:ns:loss:kge}) to Eq.~(\ref{eq:app:sub:intro:end}), we can reformulate Eq.~(\ref{eq:app:sub:replaced}) as follows:
\begin{equation}
(\ref{eq:app:sub:replaced}) \approx -\frac{1}{|D|}\sum_{(x,y) \in D} \Bigl[A(x,y)\log(\sigma(s_{\theta}(x,y)+\gamma))+\frac{1}{\nu}\sum_{y_{i}\sim p_n(y_{i}|x)}^{\nu}B(x)\log(\sigma(-s_{\theta}(x,y_i)-\gamma))\Bigr].
\label{eq:app:subsamp}
\end{equation}

\section{Experimental Details}

\begin{table*}[t]
    \centering
    \caption{Hyperparameters for each model. \textit{Batch} denotes the batch size, \textit{Dim} denotes the hidden dimension size, $\alpha$ denotes the temperature parameter for SANS, and \textit{Step} denotes the max steps in training.}
    \begin{tabular}{lllllllllllll}
    \toprule
       \multirow{2}{*}{\textbf{Model}} & \multicolumn{4}{c}{\textbf{FB15k-237}} & \multicolumn{4}{c}{\textbf{WN18RR}} & \multicolumn{4}{c}{\textbf{YAGO3-10}} \\
        \cmidrule(lr){2-5}\cmidrule(lr){6-9}\cmidrule(lr){10-13}
         & \textbf{Batch} & \textbf{Dim.} & $\alpha$ & \textbf{Steps} & \textbf{Batch} & \textbf{Dim.} & $\alpha$ & \textbf{Steps} & \textbf{Batch} & \textbf{Dim.} & $\alpha$ & \textbf{Step}\\
         \midrule
         \textbf{RESCAL} & 1024 & 500 & 1.0 & 100000 & 512 & 500 & 1.0 & 80000 & - & - & - & - \\
         \textbf{ComplEx} & 1024 & 1000 & 1.0 & 100000 & 512 & 500 & 1.0 & 80000 & - & - & - & -  \\
         \textbf{DistMult} & 1024 & 2000 & 1.0 & 100000 & 512 & 1000 & 1.0 & 80000 & - & - & - & - \\
         \textbf{TransE} & 1024 & 1000 & 1.0 & 100000 & 512 & 500 & 0.5 & 80000 & 1024 & 500 & 1.0 & 200000   \\
         \textbf{RotatE} & 1024 & 1000 & 1.0 & 100000 & 512 & 500 & 0.5 & 80000 & 1024 & 500 & 1.0 & 200000  \\
         \textbf{HAKE} & 1024 & 1000 & 1.0 & 100000 & 512 & 500 & 0.5 & 80000 & 1024 & 500 & 1.0 & 200000 \\
    \bottomrule
    \end{tabular}
    \label{tab:setting:models}
\end{table*}
\begin{table*}[t]
    \centering
    \caption{Hyperparameters for each model in \S  \ref{subsec:analysis:sans}. \textit{LR} denotes the learning rate.}
    \begin{tabular}{llllllllll}
    \toprule
    \multirow{2}{*}{\textbf{Model}} & \multicolumn{3}{c}{\textbf{FB15k-237}} & \multicolumn{3}{c}{\textbf{WN18RR}} & \multicolumn{3}{c}{\textbf{YAGO3-10}} \\
    \cmidrule(lr){2-4}\cmidrule(lr){5-7}\cmidrule(lr){8-10}
    & $\nu$ & $\gamma$ & \textbf{LR} & $\nu$ & $\gamma$ & \textbf{LR} & $\nu$ & $\gamma$ & \textbf{LR} \\
    \midrule
        \textbf{RESCAL} & 256 & 200 & 0.001 & 1024 & 200 & 0.002 & - & - & - \\
        \textbf{ComplEx} & 256 & 200 & 0.001 & 1024 & 200 & 0.002 & - & - & - \\ 
        \textbf{DistMult} & 256 & 200 & 0.001 & 1024 & 200 & 0.002 & - & - & - \\
        \textbf{TransE} & 256 & 9.0 & 0.00005 & 1024 & 6.0 & 0.00005 & 400 & 24.0 & 0.0002 \\
        \textbf{RotatE} & 256 & 9.0 & 0.00005 & 1024 & 6.0 & 0.00005 & 400 & 24.0 & 0.0002 \\
        \textbf{HAKE} & 256 & 9.0 & 0.00005 & 1024 & 6.0 & 0.00005 & 500 & 24.0 & 0.0002 \\
    \bottomrule
    \end{tabular}
    \label{tab:setting:subsampling}
\end{table*}

\subsection{Detailed Hyperparameters for \S \ref{subsec:analysis:margin}}
\label{app:analysis:margin:details}

For each model, we used the hyperparameters of FB15k-237 and WN18RR in Table \ref{tab:setting:models}. We explained other hyperparameters and settings in \S \ref{subsec:analysis:margin}.

\subsection{Detailed Hyperparameters for \S \ref{subsec:analysis:nu}}
\label{app:analysis:nu:details}

For each model, we used the hyperparameters of FB15k-237 and WN18RR in Table \ref{tab:setting:models}. We explained other hyperparameters and settings in \S \ref{subsec:analysis:nu}.

\subsection{Detailed Hyperparameters for \S \ref{subsec:analysis:nu-margin}}
\label{app:analysis:nu-margin:details}

For each model, we used the hyperparameters of FB15k-237 and WN18RR in Table \ref{tab:setting:models}. We explained other hyperparameters and settings in \S \ref{subsec:analysis:nu-margin}.

\subsection{Detailed Hyperparameters for \S \ref{subsec:analysis:sans}}
\label{app:analysis:sans:details}

For each model, we used the hyperparameters listed in Tables \ref{tab:setting:models} and \ref{tab:setting:subsampling}. We described other settings in \S \ref{subsec:analysis:sans}.

\end{document}